\documentclass[10pt,twocolumn,letterpaper]{article}

\usepackage{iccv}
\usepackage{times}
\usepackage{epsfig}
\usepackage{graphicx}
\usepackage{amsmath}
\usepackage{amssymb}
\usepackage{bm}
\usepackage{algorithm}
\usepackage{algorithmicx}
\usepackage[noend]{algpseudocode}
\usepackage{cuted}
\usepackage{capt-of}

\usepackage[breaklinks=true,bookmarks=false]{hyperref}

\iccvfinalcopy % *** Uncomment this line for the final submission

\ificcvfinal\fi

\begin{document}

%%%%%%%%% TITLE
\title{Photo-Realistic Facial Details Synthesis From Single Image}

\author{Anpei Chen$^{1}$\\
\and
Zhang Chen$^{1}$\\
\and
Guli Zhang$^1$\\
\and
Ziheng Zhang$^1$\\
\and
Kenny Mitchell$^{2}$ \qquad \qquad Jingyi Yu$^1$\\
$^{1}$ ShanghaiTech University \qquad \qquad $^{2}$ Edinburgh\ Napier\ University\\
{\tt\small {\{chenap,chenzhang,zhanggl,yujingyi\}}@shanghaitech.edu.cn \qquad K.Mitchell2@napier.ac.uk}
}
\maketitle
\ificcvfinal\thispagestyle{empty}\fi

\begin{strip}\centering
\includegraphics[width=\textwidth]{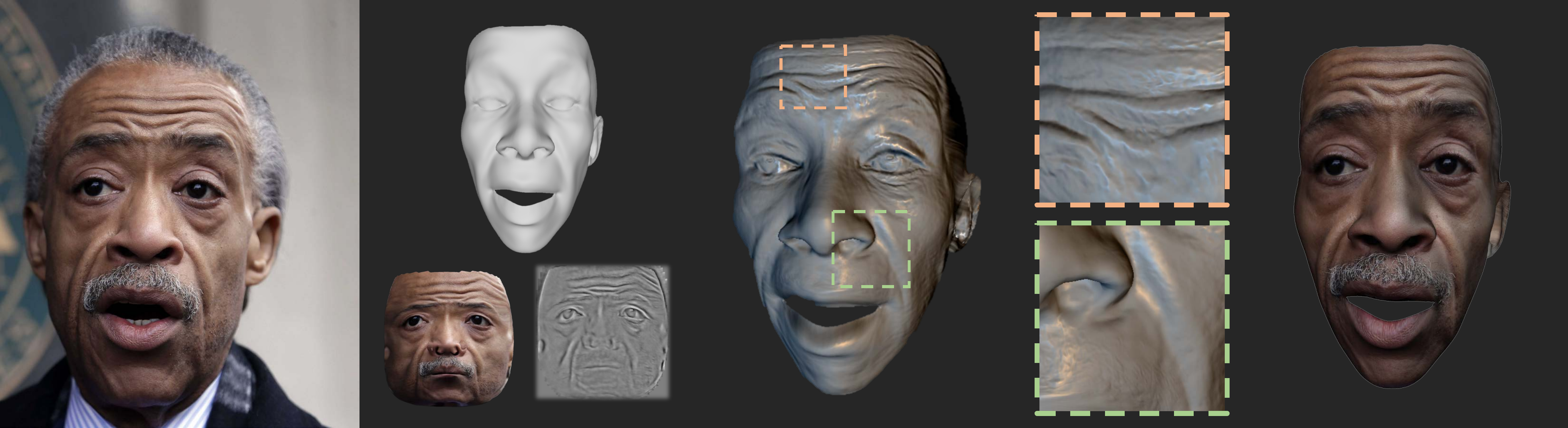}
\captionof{figure}{From left to right: input face image; proxy 3D face, texture and displacement map produced by our framework; detailed face geometry with estimated displacement map applied on the proxy 3D face; and re-rendered facial image.}
\label{fig:teaser}
\end{strip}

%%%%%%%%% ABSTRACT
\begin{abstract}
%   We present a single-image 3D face synthesis technique that can handle challenging facial expressions and lighting. Our technique, called Emotion-to-Face (E2F), employs emotion estimation for proxy geometry generation and combines supervised and unsupervised learning for detail synthesis using both geometry and images for training. On proxy generation, we conduct emotion prediction and then use the result to determine an expression informed proxy. On detail synthesis, we present a Deep Facial Detail Net (DFDN) based on Conditional Generative Adversarial Net (\textsl{CGAN}) that employs both geometry and appearance loss functions. For geometry, we capture 366 high-quality 3D scans from 122 different subjects under 3 facial expressions. For appearance, we use additional 20K in-the-wild face images and apply image-based rendering to accommodate lighting variations. Comprehensive experiments demonstrate that our E2F framework can produce high-quality 3D faces with very realistic details from images under complex lighting and with challenging facial expressions. 
   We present a single-image 3D face synthesis technique that can handle challenging facial expressions while recovering fine geometric details. Our technique employs expression analysis for proxy face geometry generation and combines supervised and unsupervised learning for facial detail synthesis. On proxy generation, we conduct emotion prediction to determine a new expression-informed proxy. On detail synthesis, we present a Deep Facial Detail Net (DFDN) based on Conditional Generative Adversarial Net (CGAN) that employs both geometry and appearance loss functions. For geometry, we capture 366 high-quality 3D scans from 122 different subjects under 3 facial expressions. For appearance, we use additional 163K in-the-wild face images and apply image-based rendering to accommodate lighting variations. Comprehensive experiments demonstrate that our framework can produce high-quality 3D faces with realistic details under challenging facial expressions. 
\end{abstract}
\ificcvfinal\pagestyle{empty}\fi

%%%%%%%%% BODY TEXT
\section{Introduction}

Producing high quality human faces with fine geometric details has been a core research area in computer vision and graphics. Geometric structure details such as wrinkles are important indicators of age and facial expression, and are essential for producing realistic virtual human \cite{alexander2010digital}. Successful solutions by far rely on complex and often expensive capture systems such as stereo-based camera domes \cite{graham2015near} or photometric-based LightStage \cite{ma2007rapid,cao2018sparse}. Although such solutions have become increasingly popular and affordable with the availability of low-cost cameras and lights, they are still bulky and hence do not support portable scanning. In addition, they are vulnerable to low texture regions such as bare skins. 

% An increasing number of mobile devices are also equipped with 3D face scanning capabilities by employing structured light 3D sensors (e.g., the iPhone X series), stereo cameras (e.g., Huawei P20), or time-of-flight cameras (e.g., OPPO R17 Pro). The scanned 3D faces, combined with aligned images, can serve as useful biometrics alternatives to fingerprints. However, 3D face geometry obtained on mobile devices is noisy, low-resolution, and even incomplete, largely attributable to low resolution depth sensors: 640x360 on iPhone X, 240x180 on OPPO, etc. Stereo based solutions can potentially produce higher resolution reconstruction but its small camera baseline (e.g., less than 12 mm on Huawei P20) prohibits the sensor from capturing millimeter-level geometric details.

We aim to produce high-quality 3D faces with fine geometric details from a single image, with quality comparable to those produced from the dome systems and LightStage. Existing single-image solutions first construct a 3D proxy face from templates and then refine the proxy by deforming geometry and adding details. Such proxies can be derived from 3D Morphable Model (3DMM) \cite{cao20133d,cao2014displaced,saito2016real,thies2016face2face,garrido2016reconstruction} by blending base face geometry. More complex techniques employ sparse coding on 3D face dictionaries to further improve robustness and quality \cite{romdhani2005estimating,cao20133d,cao2014displaced,saito2016real,thies2016face2face,garrido2016reconstruction,hu2017efficient,schonborn2017markov,booth20183d}. However, artifacts arise from these approaches such as over-smoothing and incorrect expression, where a relatively small number of parameters are used to approximate the high dimensional space for real face. Shape-from-shading \cite{kemelmacher20113d}, photometric stereo \cite{cao2018sparse}, and deep learning \cite{tran2017regressing,richardson20163d,dou2017end} have been used to generate the missing details. However, existing methods have limits in attaining correct shape under unseen emotional expressions and lighting, thus delivering insufficient or inaccurate geometric details, as shown in Fig. \ref{fig:closeup} 

In this paper, we present a novel learning-based technique to produce accurate geometric details from a single face image. Our approach takes into account emotion, expression and appearance. For proxy generation, we employ the \emph{Basel Face Model} (\emph{BFM}) \cite{gerig2018morphable} composed of shape, expression and surface reflectance (albedo). 3D expressions, however, exhibit strong ambiguity after being projected onto 2D images: a pair of 3D meshes that represent very different emotional expressions can have similar 2D landmarks on images. Therefore, we first devise a learning-based approach to conduct emotion prediction and then use the result to determine an expression-informed proxy.
% (section \ref{E2F}).
%coupled with an iterative two-stage supervised geometry and unsupervised photometric refinement process. Comprehensive experiments demonstrate that our technique significantly outperforms the state-of-the-art and produces high quality 3D face geometry with unprecedented realism. 

% (section \ref{E2F}).

For geometric detail synthesis, we devise a \textsl{Deep Facial Detail Net} (\textsl{DFDN}) based on Conditional Generative Adversarial Net (CGAN) to map an image patch to a detailed displacement map. Our \textsl{DFDN} has two components: a medium scale geometry module that learns the PCA coefficients (in our case 64) of each patch and a fine scale geometry module that refines the PCA-based result with additional details. For training, we captured a total of 366 high quality 3D scans from 122 different subjects under three facial expressions (one neutral and two extreme expressions). We augment the training data with 340 high resolution meshes from \textsl{ICT-3DRFE} \cite{stratou2011effect}. The loss function is defined in terms of geometric differences between the estimation and the ground truth. However, we observe that these training data are still insufficient to cover a wide range of lighting conditions. Hence, we introduce an additional unsupervised learning procedure (with an additional 163K images captured in the wild) where for each image we obtain its proxy geometry using our emotion-driven shape estimator and then approximate the corresponding environment lighting using spherical harmonics (SH). We use \textsl{DFDN} to obtain an estimate of the geometry, but since we do not have the ground truth geometry, we re-render these results using the estimated albedo and environment lighting, and compute the loss function in terms of the image differences. Finally, we alternate the supervised and the unsupervised learning processes, on geometry and image, respectively. We have released our code, pre-trained models and results\footnote{\url{https://github.com/apchenstu/Facial_Details_Synthesis.git}}.

\begin{figure*}[t]
\centering
\includegraphics[width=0.85\textwidth]{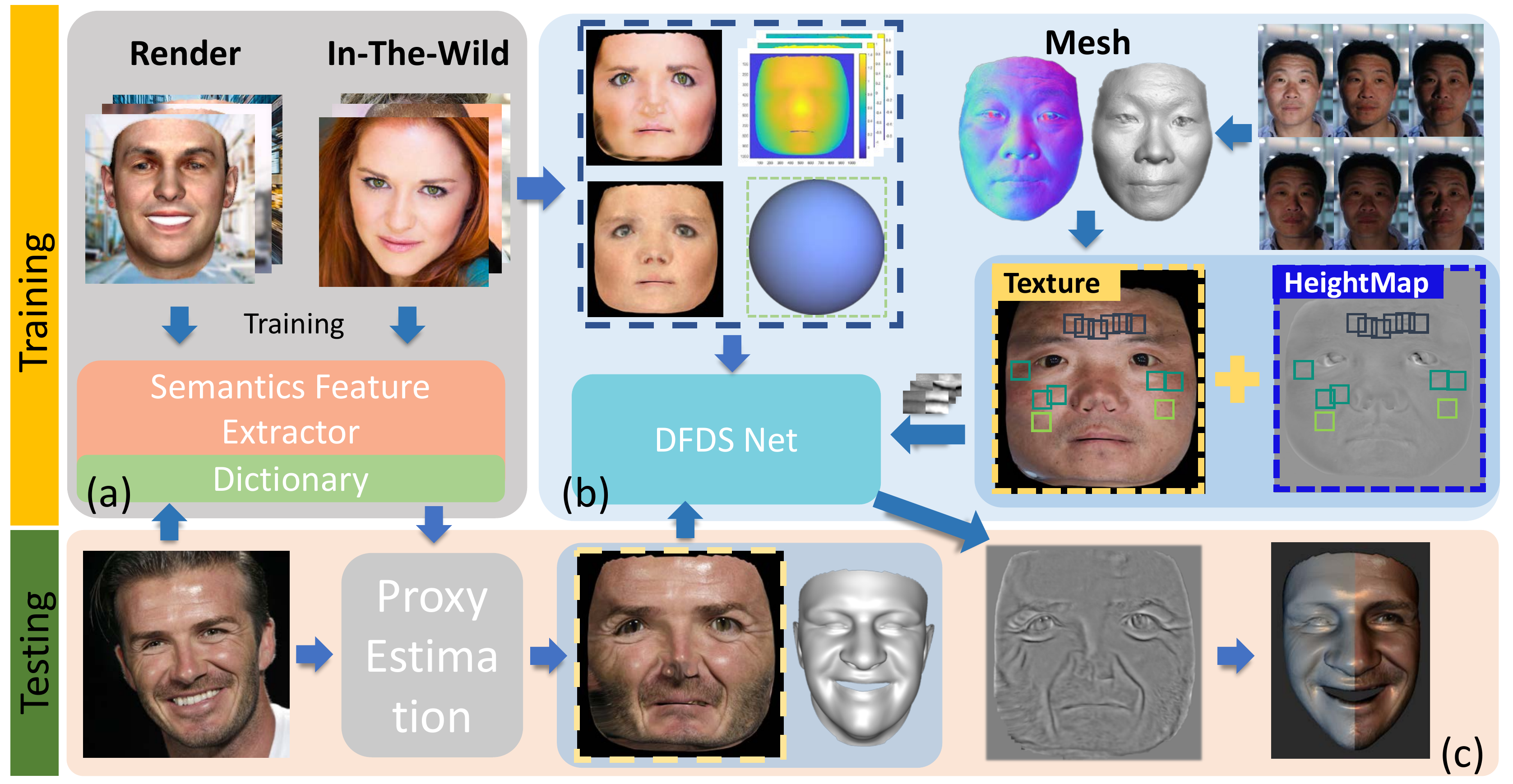}
\caption{Our processing pipeline. Top: training stage for (a) emotion-driven proxy generation and (b) facial detail synthesis. Bottom: testing stage for an input image.}
\label{fig:pipeline}
\end{figure*}

%-------------------------------------------------------------------------
\section{Related Work}
Existing approaches for producing high quality 3D face geometry either rely on reconstruction or synthesis. 

\textbf{Reconstruction-based Techniques.} Multi-View Stereo (MVS) 3D face reconstruction systems employ stereo \cite{matas2004robust} or structure-from-motion \cite{westoby2012structure}. A sparse set of cameras produce large scale geometry \cite{graham2015near} whereas denser and hence more expensive settings \cite{alexander2010digital} provide more accurate measurements. In either case, the reconstruction quality depends heavily on the feature matching results as they act as anchor points dominating the final shape. For regions with few textures such as bare skin, the reconstruction tends to be overly smooth due to lack of features. For example, wrinkles caused by facial expressions are particularly difficult to reconstruct: even though they cause shading variations, their geometry is too slight to capture using stereo, especially when the camera baseline is small. 
% Valgaerts et al. \cite{valgaerts2012lightweight} combined image-based tracking with shading-based geometry refinement from stereo image sequences. 
Recently, Graham et al. \cite{graham2015near} use 24 entry-level DSLR photogrammetry cameras and 6 ring flashes to capture facial specular response independently and then combine shape-from-chroma and shape-from-specularity for high quality reconstruction. 
	
Another class of multi-shot techniques employed in face reconstruction is Photometric Stereo (PS). PS is based on analyzing image intensity variations under different illuminations from a fixed viewpoint. Instead of directly reconstructing 3D geometry, PS intends to first recover the normal map and then the 3D mesh, e.g., via normal integration. 
% The most notable example is the USC Light Stage. 
A common artifact in PS is low-frequency distortions in the final reconstruction \cite{papadhimitri2013new, tankus2005photometric} caused by perspective projection violating the orthographic assumption. Accurate calibrations on both the light sources and camera, though able to mitigate the problem, are cumbersome. Most recent techniques \cite{park2017robust, wu2011fusing, esteban2008multiview} combine PS with MVS by using the MVS results as a proxy for calibration and then refine the results. Aliaga et al. \cite{aliaga2010self} simulates a MVS setup by employing multiple digital projectors as both light sources and virtual cameras. 
% It is also possible to combine PS with active 3D sensing such as structured light based or time-of-flight based RGBD sensors \cite{haque2014high, zhang2012edge}. 
We refer the readers to \cite{ackermann2015survey} for a comprehensive review of PS variants.

\textbf{Synthesis-based approaches.} The availability of high quality mobile cameras and the demand on portable 3D scanning have promoted significant advances on producing high quality 3D faces from a single image. The seminal work of Blanz and Vetter \cite{blanz1999morphable} pre-captures a database of face models and extracts a 3D morphable model (3DMM) composed of base shapes and albedos. Given an input image, it finds the optimal combination of the base models to fit the input. Their technique can also handle geometric deformations under expressions \cite{paysan20093d,gerig2018morphable} if the database includes expressions, e.g., captured by RGBD cameras \cite{cao2014facewarehouse}. More extensive facial databases have been recently made publicly available \cite{yin20063d,huber2016multiresolution,vlasic2005face,li2017learning,booth2018large}, with an emphasis on handling complex expressions \cite{li2017learning,booth2018large}. Most recently, Li \textit{et al.} \cite{li2017learning} capture pose and articulations of jaw, neck, and eyeballs with over 33,000 3D scans that have helped boost the performance of single-image/video face reconstruction/tracking \cite{romdhani2005estimating,cao20133d,cao2014displaced,shi2014automatic,zhu2015discriminative,saito2016real,thies2016face2face,garrido2016reconstruction,hu2017efficient,schonborn2017markov,booth20183d}. The current databases, however, still lack mid- and high-frequency geometric details such as wrinkles and pores that are epitomes to realistic 3D faces. Shading based compensations can improve the visual appearance \cite{garrido2016reconstruction,jiang20183d} but remain far behind quality reconstruction of photos. 

 Our approach is part of the latest endeavor that uses learning to recover 3D proxy and synthesize fine geometric details from a single image. For proxy generation, real \cite{tran2017extreme} or synthetically rendered \cite{richardson20163d,dou2017end} face images are used as training datasets, and convolutional neural networks (CNNs) are then used to estimate 3D model parameters. Zhu \textit{et al.} \cite{zhu2016face} use synthesized images in profile views to enable accurate proxy alignment for large poses. Chang \textit{et al.} \cite{chang2018expnet} bypass landmark detection to better regress for expression parameters. Tewari \textit{et al.} \cite{tewari2017mofa} adopt a self-supervised approach based on an autoencoder where a novel decoder depicts the image formation process. Kim \textit{et al.} \cite{kim2018inversefacenet} combine the advantages of both synthetic and real data to \cite{tewari2018self} jointly learn a parametric face model and a regressor for its corresponding parameters. However, these methods do not exploit emotion information and cannot fully recover expression traits.
 For detail synthesis, Sela \textit{et al.} \cite{sela2017unrestricted} use synthetic images for training but directly recover depth and correspondence maps instead of model parameters. Richardson \textit{et al.} \cite{richardson2017learning} apply supervised learning to first recover model parameters and then employ shape-from-shading (SfS) to recover fine details. Li \textit{et al.} \cite{li2018face} incorporate SfS with albedo prior masks and a depth-image gradients constraint to better preserve facial details. Guo \textit{et al.} \cite{guo2018cnn} adopt a two-stage network to reconstruct facial geometry at different scales. Tran \textit{et al.} \cite{tran2018extreme} represent details as bump map and further handle occlusion by hole filling. Learning based techniques can also produce volumetric representations \cite{jackson2017large} or normal fields \cite{sengupta2018sfsnet}. Yet, few approaches can generate very fine geometric details. Cao \textit{et.al} \cite{cao2015real} capture 18 high quality scans and employ a principal component analysis (PCA) model to emulate wrinkles as displacement maps. Huynh \textit{et al.} \cite{huynh2018mesoscopic} use high precision 3D scans from the LightStage \cite{ghosh2011multiview}. Although effective, their technique assumes similar environment lighting as the LightStage. 

\section{Expression-Aware Proxy Generation}
\label{E2F}
Our first step is to obtain a proxy 3D face with accurate facial expressions. We employ the \emph{Basel Face Model} (\emph{BFM}) \cite{gerig2018morphable}, which consists of three components: shape $M_{\emph{sha}}$, expression $M_{\emph{exp}}$ and albedo $M_{\emph{alb}}$. Shape $M_{\emph{sha}}$ and expression $M_{\emph{exp}}$ determine vertex positions while albedo $M_{\emph{alb}}$ encodes per-vertex albedo:
\begin{eqnarray}
M_{\emph{sha}}(\bm{\alpha}) &= \bm{a}_{sha}+ \mathbf{E}_{sha}\cdot \bm{\alpha} \\
M_{\emph{exp}}(\bm{\beta}) &= \bm{a}_{exp}+ \mathbf{E}_{exp}\cdot \bm{\beta} \\
M_{\emph{alb}}(\bm{\gamma}) &= \bm{a}_{alb}+ \mathbf{E}_{alb}\cdot \bm{\gamma}
\end{eqnarray}

where $\bm{a}_{sha},\bm{a}_{exp},\bm{a}_{alb}\in\mathbb{R}^{3n}$ represent the mean of corresponding PCA space. $\mathbf{E}_{sha}\in\mathbb{R}^{3n \times 199}$, $\mathbf{E}_{exp} \in \mathbb{R}^{3n \times 100}$ contain basis vectors for shape and expression while $\mathbf{E}_{alb} \in \mathbb{R}^{3n \times 199}$ contain basis vectors for albedo. $\bm{\alpha}, \bm{\beta} , \bm{\gamma}$ correspond to the parameters of the PCA model.

\subsection{Proxy Estimation}
Given a 2D image, we first extract 2D facial landmarks $\mathbf{L}\in\mathbb{R}^{2m}$ and use the results to compute PCA parameters $\bm{\alpha}$, $\bm{\beta}$ for estimation of proxy shape. Specifically, we set out to find the parameters that minimize the reprojection error on landmarks:
\begin{align}\label{eq:proxy shape}
E = \sum_{k} w_{k} \Arrowvert \mathbf{L}_k - P(\mathbf{l}_k(\bm{\alpha}, \bm{\beta})) \Arrowvert_2 + \lambda_s \Arrowvert \bm{\alpha} \Arrowvert _2%\\
% \lambda_e \Arrowvert \beta-\beta_{prior}\Arrowvert _2 
\end{align}
where $\mathbf{l}_k(\bm{\alpha}, \bm{\beta})$ corresponds to the $k$th facial vertex landmark and $P(\cdot)$ is the camera projection operator that maps 3D vertices to 2D image coordinates. $w_{k}$ controls the weight for each facial landmark whereas $\lambda_s$ imposes regularization on the shape parameters. %$\beta_{prior}$ denotes the emotion-based expression prior that we will describe in Section~\ref{sec:emotion}.

To solve for Eq.~\ref{eq:proxy shape}, we use the iterative linear method in ~\cite{huber2016multiresolution}. Specifically, the camera projection operator $P(\cdot)$ is parameterized as an affine camera matrix. For expression parameters $\bm{\beta}$, different from~\cite{huber2016multiresolution}, we fix it as prior parameters $\bm{\beta}_{prior}$ computed in Section~\ref{sec:emotion}. During each round of iterations, we first fix $\bm{\alpha}$ and solve for $P(\cdot)$ using the \emph{Gold Standard Algorithm}~\cite{hartley2003multiple}. We then fix $P(\cdot)$ and solve for $\bm{\alpha}$. To bootstrap this iterative scheme, we initialize $\bm{\alpha}$ as $\mathbf{0}$.

\subsection{Imposing Expression as Priors}\label{sec:emotion}
\begin{figure}[t]
\begin{center}
   \includegraphics[width=0.9\linewidth]{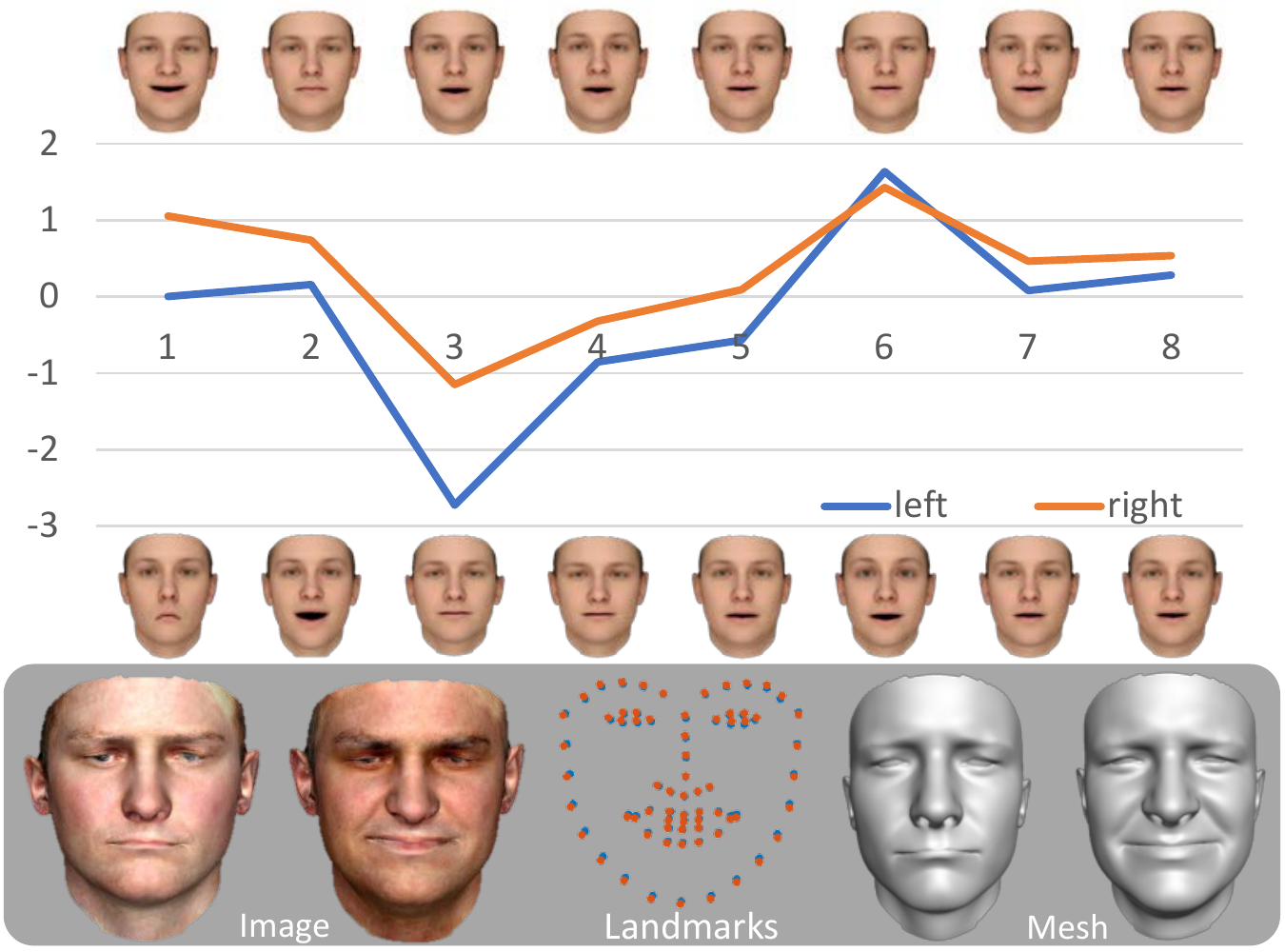}
\end{center}
   \caption{Expression projection ambiguity. Top: Visualization of the two models' first eight dimensions of 3D expression parameters. Bottom: Rendered 2D facial images, their landmarks layered onto each other, and their corresponding meshes. `left, right' refer to both rendered images and meshes.}
\label{fig:expression ambiguity}
\end{figure}

The most challenging component in proxy estimation is expression. 3D expressions exhibit a significant ambiguity after being projected onto 2D images, e.g., different expressions may have similar 2D facial landmarks after projection. Fig.~\ref{fig:expression ambiguity} shows an example of this ambiguity: the landmarks of the two faces are extremely close to each other while their expression parameters and shapes are vastly different, especially around nasolabial folds. So it is hard to define or train a mapping directly from image to 3D expression. In our experiments, we also observe that the reprojection-based loss function can easily fall into local minimum that reflects such ambiguity. 

% Having observed that facial expression is strongly correlated with human emotion as people tend to express their emotions through facial expression changes, 

We propose to use facial semantic information to narrow the proxy parameter solving space via converting the problem into a conditional distribution. Our high level semantic features comprise of emotion features and physically-based appearance features (e.g. FACS).

To obtain emotion features, we reuse $\textsl{AffectNet}$ dataset~\cite{mollahosseini2017affectnet} to train an emotion feature predictor \textsl{Emotion-Net}. The dataset contains 11 discrete emotion categories and about 450K annotated images. We utilize the "sequential fully-CNN" \cite{arriaga2017real} architecture to train our emotion feature predictor and use the output of the second last layer $f \in \mathbb{R}^{128}$ as the feature vector to represent human emotions. More details could be found in our released code. Next, we randomly generate expression parameters $\bm{\beta}$ from normal distribution in interval $[-3, 3]$ and render 90K images with different facial expressions. We feed the images into the trained \textsl{Emotion-Net} and obtain a total of 90K emotion feature vectors. We also use ~\cite{baltruvsaitis2015cross} to estimate 90K appearance feature vectors on the images. Concatenating these emotion feature vectors along with their corresponding appearance feature vectors, we obtain the semantic feature vector for each of the 90K images. We then formulate a dictionary $\Psi\colon\Psi_{sem}\to\Psi_{exp}$ that record the mapping from semantic features $\Psi_{sem}$ to expression parameters $\Psi_{exp}$. Once we obtain the trained model and the expression dictionary, we can predict expression parameters $\bm{\beta}_{prior}$ as a prior for proxy estimation. 

Given a new image $I$, we first feed it to \textsl{Emotion-Net} and appearance feature predictor to obtain its semantic feature vector. We then find its closest semantic feature vector in the dictionary and use the corresponding expression parameters for $\bm{\beta}_{prior}$:
\begin{equation}
\label{eq:refine expression}
     \bm{\beta}_{prior} = \Psi (\mathop{\arg\min}_{\psi_{sem}} \Arrowvert\textsl{Emotion-Net}(I)-\psi_{sem}\Arrowvert_2)
\end{equation}
%Then we apply expression parameters $\beta_{prior}$ to conduct a proxy mesh estimation.
% and conduct camera intrinsic/extrinsic estimation, the strict forward approach is  direct iterate solving the minimize the 3D landmarks re-projection errors:
% \begin{equation}
% E = \sum_{j=1}^{68} w_{j}(\Arrowvert P(I,L_j) - l_{j} \Arrowvert ^2_2)
% \end{equation}
%Also our detail caving pipeline do not limit to a specially proxy mesh reconstruction algorithms, i.e. mobile depth camera,  structure light et al. However, which contain a lot unknown factors like mesh scale, orientation, texture's UV mapping so on.  Our DFDN learn details from facial texture information, it is hard to handle any type input data, it is necessary to align the proxy mesh and texture to a specified common coordinate system, we also provide a solution to those input data, shown in supplementary materials.

%Although more complicated schemes such as sparse reconstruction can be used, in our experiments we find that the nearest one is sufficiently robust. Our approach guarantees that each image maps to a unique expression and resolves the ambiguity caused by reprojection-based landmark matching. 

%We validate this method is very effective and robust,as shown in Fig.XXX.

\begin{figure*}[t]
\begin{center}
   \includegraphics[width=0.85\textwidth]{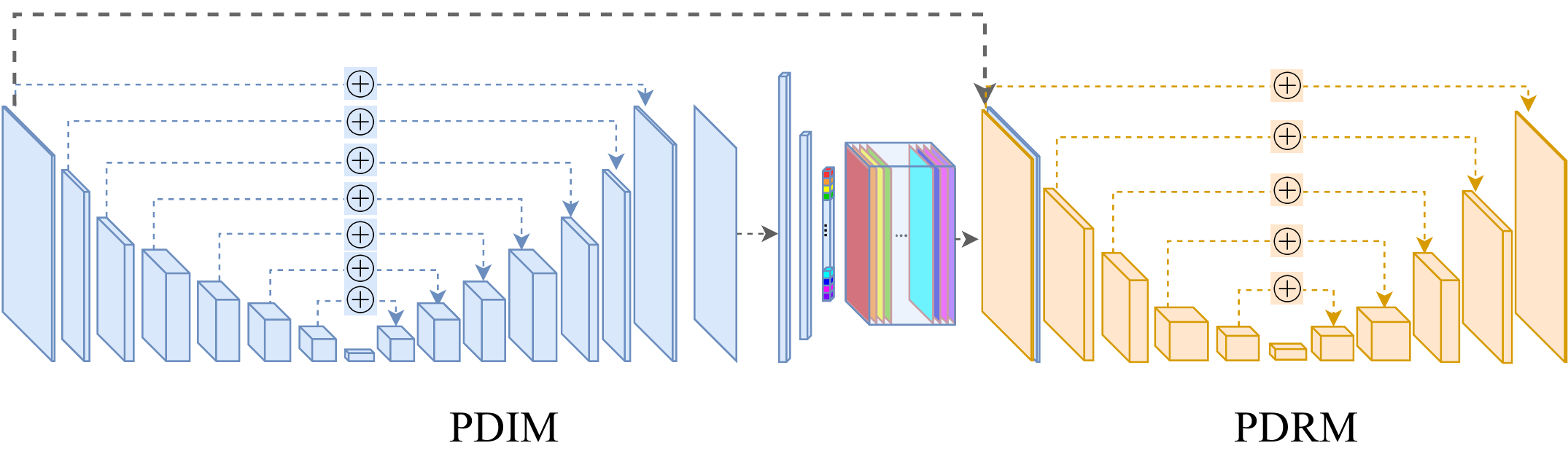}
\end{center}
   \caption{Network architecture for facial detail synthesis. PDIM for medium-frequency detail (wrinkles) synthesis and PDRM for high-frequency detail (pores) synthesis.}
\label{fig:network}
\end{figure*}

\section{Deep Facial Detail Synthesis}
\label{DFDN}
With the 3D proxy face, we synthesize geometric details by estimating displacement map and applying to the proxy mesh. The key observation here is that for facial details such as wrinkles, there is strong correlation between geometry and appearance. 

\subsection{Network Architecture}
Fig.~\ref{fig:network} shows our \textsl{Deep Facial Detail Net} (\textsl{DFDN}) with two main cascaded modules. The \textsl{Partial Detail Inference Module} (\textsl{PDIM}) takes 2D image patches as inputs and generates 3D facial geometric details using a PCA-based technique (Section \ref{subsec:gepmetric loss}). Such a scheme dramatically reduces the parameter space and is stable for both training and inference process. However, PCA-based approximations lose high-frequency features that are critical to fine detail synthesis. We therefore introduce the \textsl{Partial Detail Refinement Module} (\textsl{PDRM}) to further refine high-frequency details. The reason we explicitly break down facial inference procedure into linear approximation and non-linear refinement is that facial details consist of both regular patterns like wrinkles and characteristic features such as pores and spots. By using a two-step scheme, we encode such priors into our network.

In \textsl{PDIM} module, we use \textsl{UNet-8} structure concatenated with 4 fully connected layers to learn the mapping from texture map to PCA representation of displacement map. The sizes of the 4 fully connected layers are $2048,1024,512,64$. Except for the last fully connected layer, each linear layer is followed by an \textsl{ReLU} activation layer. In the subsequent \textsl{PDRM} module, we use \textsl{UNet-6}, i.e. 6 layers of convolution and deconvolution, each of which uses 4$\times$4 kernel, 2 for stride size, and 1 for padding size. Apart from this, we adopt \textsl{LeakReLU} activation layer except for the last convolution layer and then employ \textsl{tanh} activation.

To train \textsl{PDIM} and \textsl{PDRM} modules, we combine supervised and unsupervised training techniques based on Conditional Generative Adversarial Nets (CGAN), aiming to handle variations in facial texture, illumination, pose and expression. Specifically, we collect 706 high precision 3D human faces and over 163K unlabeled facial images captured in-the-wild to learn a mapping from the observed image $x$ and the random noise vector $z$ to the target displacement map $y$ by minimizing the generator objective $G$ and maximizing log-probability of 'fooling' discriminator $D$ as: 
% arg\underset{G}{min}\ \underset{D}{max}\
\begin{equation}
\begin{aligned}
% \mathcal{L} &= \mathop{\arg\min}_{G}\max_{D} (\mathcal{L}_{cGAN(G,D)}+\lambda \mathcal{L}_{L1(G)}), 
\min_{G}\max_{D} (\mathcal{L}_{cGAN(G,D)}+\lambda \mathcal{L}_{L1(G)}), 
\end{aligned}
\end{equation}
where we set $\lambda=100$ in all our experiments and
%\begin{center}
%\begin{equations}
\begin{align}
\mathcal{L}_{cGAN(G,D)} = &\mathbb{E}_{x,y}[log\ D(x,y)]+ \nonumber\\
                          &\mathbb{E}_{x,z}[log(1-D(x,G(x,z)))]. 
\end{align}

%\end{equations}
%\end{center}
A major drawback of the supervised learning scheme mentioned above is that the training data, captured under fixed setting (controlled lighting, expression, etc.), are insufficient to emulate real face images that exhibit strong variations caused by environment lighting and expressions. We hence devise a semi-supervised generator $G$, exploiting labeled 3D face scans for supervised loss $\mathcal{L}_{scans}$ as well as image-based modeling and rendering for unsupervised reconstruction loss $\mathcal{L}_{recon}$ as:
\begin{equation}
\mathcal{L}_{L1(G)} = \mathcal{L}_{scans}(x,z,y)+\eta \mathcal{L}_{recon}(x),
\end{equation}
where $x$ is input image, $z$ is random noise vector and $y$ is groundtruth displacement map. $\eta$ controls the contribution of reconstruction loss and we fix it as $0.5$ in our case. In the following subsections, we discuss how to construct the supervised loss $\mathcal{L}_{scans}$ for geometry and unsupervised loss $\mathcal{L}_{recon}$ for appearance.

\subsection{Geometry Loss}
\label{subsec:gepmetric loss}
The geometry loss compares the estimated displacement map with ground truth. To do so, we need to capture ground truth facial geometry with fine details.

\textbf{Face Scan Capture.} To acquire training datasets, we implement a small-scale facial capture system similar to \cite{cao2018sparse} and further enhance photometric stereo with multi-view stereo: the former can produce high quality local details but is subject to global deformation whereas the latter shows good performance on low frequency geometry and can effectively correct deformation.

Our capture system contains 5 Cannon 760D DSLRs and 9 polarized flash lights. We capture a total of 23 images for each scan, with uniform illumination from 5 different viewpoints and 9 pairs of vertically polarized lighting images (only from the central viewpoint). The complete acquisition process only lasts about two seconds. For mesh reconstruction, we first apply multi-view reconstruction on the 5 images with uniform illumination. We then extract the specular/diffuse components from the remaining image pairs and calculate diffuse/specular normal maps respectively using photometric stereo. The multi-view stereo results serve as a depth prior $z_0$ for normal integration \cite{queau2015reconstruction} in photometric stereo as: 
\begin{equation}
\begin{aligned}
 \min   \iint_{(u,v)\in I} [(\nabla z(u,v)-[p(u,v),q(u,v)]^\top )^2 \\
                             + \mu (z(u,v)-z_{0}(u,v))^2] du dv, 
\end{aligned}
\end{equation}
where $u,v$ represents image coordinates, $p,q$ represents approximations to $\partial_u z$ and $\partial_v z$ respectively, $z_{0}$ is the depth prior. $\mu$ controls the contribution of prior depth $z_0$. In order to generate geometry pairs with and without details, we set the weight parameter $\mu$ to $1e^{-5}$ and $1e^{-3}$ respectively. Then we obtain a ground truth displacement map for each geometry pair.

\textbf{PCA-Based Displacement Map.} In our training scheme, we choose not to directly feed complete face images as inputs to the network: such training can easily cause overfitting since we do not have sufficient 3D face models with fine details to start with. Instead, we observe that despite large scale variations on different faces, local texture details present strong similarities even if the faces appear vastly different. Hence we adopt the idea from \cite{cao2015real,cao2018sparse} to enhance our network generalization by training the network with texture/displacement patches of $256\times256$ resolution. We model the displacement using PCA, where each patch is a linear combination of 64 basis patches. 

Our geometric loss is then defined as:
\begin{align}
\mathcal{L}_{scans}(x,z,y) & = \sum||\mathcal{PCA}(\mathcal{PDIM}(x,z)) - y||_1 + \nonumber\\
&||\mathcal{PDRM}(\mathcal{PCA}(\mathcal{PDIM}(x,z)))-y||_1,
\end{align}
where $\mathcal{PCA}(\cdot)$ uses input PCA coefficients to linearly combine basis patches. By using the geometry loss, we combine the loss in PCA space with the per-pixel loss to recover finer details.

For patch sampling, we unfold each facial image into a $2048\times2048$ resolution texture map and regionally sample training patches based on semantic facial segmentation. For the sampling scheme, we iteratively reduce the displacement map gradient with a weighted Gaussian kernel for training set while we uniformly sample patches with 50\% overlap during inference.

\begin{figure}[t]\label{fig:albedo_illumination}
\begin{center}
   \includegraphics[width=1\linewidth]{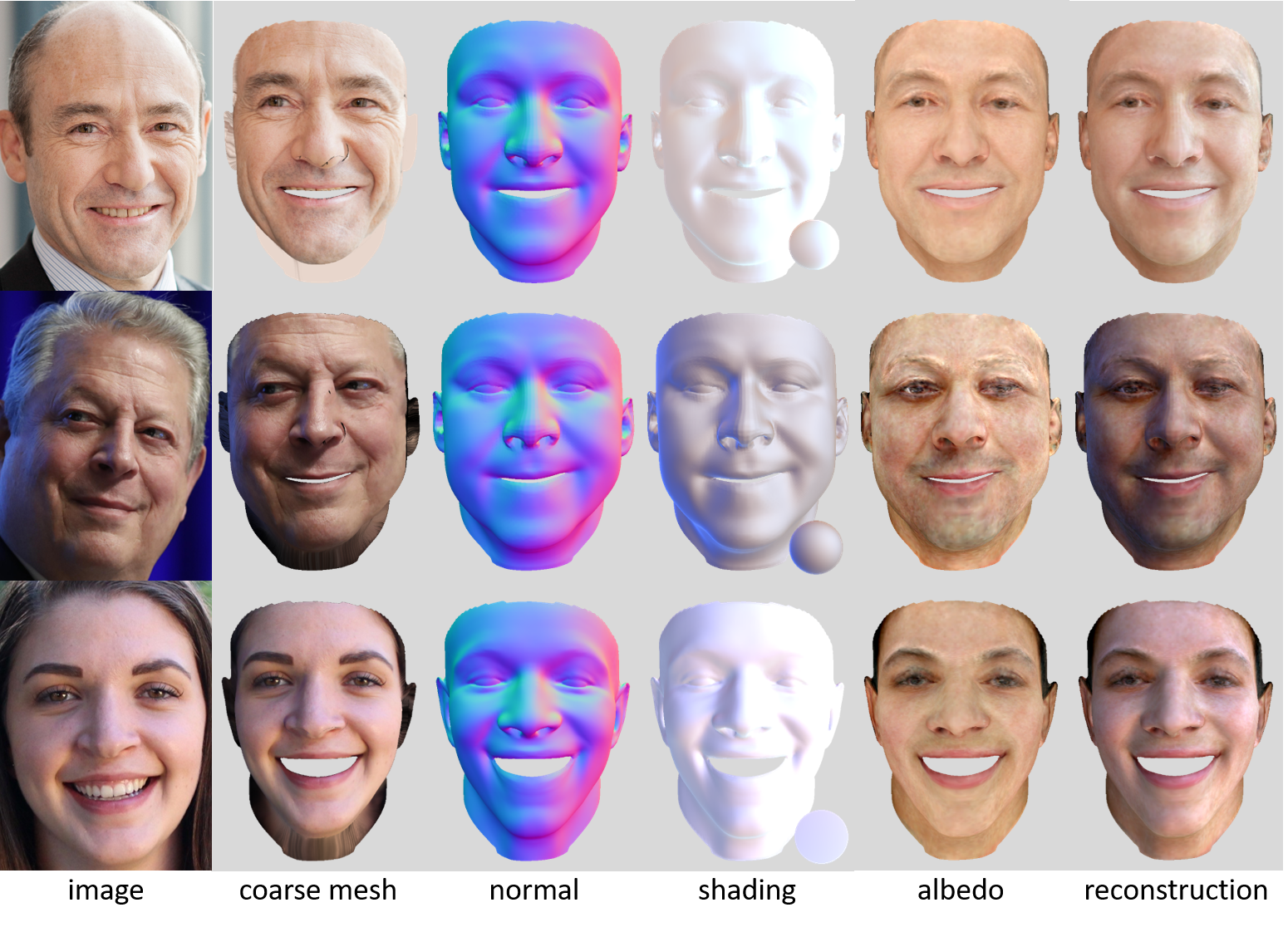}
\end{center}
   \caption{From left to right: from input image, we estimate proxy mesh, normal, lighting/shading, and albedo to re-render an image.}
\label{fig:albedo}
\end{figure}

\subsection{Appearance Loss}
Recall that the small amount of labeled facial geometry is insufficient to cover a broad range of illumination conditions and surface reflectance. Thus, we further adopt a rendering-based, unsupervised learning approach: we obtain 163K in-the-wild images, estimate its proxy (using the approach in Section 3.2) and geometric details (using \textsl{DFDN}), and then use this information to calculate lighting and albedo. Finally, we re-render an image with all these estimations and compute reconstruction loss against the input image during training.

To obtain per-pixel normals with geometric details added, we propose an texture space manipulation using the proxy mesh's position map $\mathcal{P}_{proxy}$ (shown in Fig.~\ref{fig:pipeline}, the middle of first row) and the output displacement map $G(u,v)$ from \textsl{DFDN}:

\begin{equation}
\mathcal{P}_{fine}(u,v) = \mathcal{P}_{proxy}(u,v) + G(u,v) \cdot \mathcal{N}_{proxy}(u,v)
\end{equation}
\begin{equation}
\mathcal{N}_{fine} = \mathcal{F}(\mathcal{P}_{fine})
\end{equation}

where $\mathcal{N}_{proxy}$, $\mathcal{N}_{fine}$ represent normal map of proxy and fine scale geometry, and $\mathcal{P}_{fine}$ is the position map of detailed mesh. $\mathcal{N}_{proxy}$, $\mathcal{P}_{proxy}$ are pre-rendered by a traditional rasterization rendering pipeline. %as shown in the top row of Fig.\ref{fig:pipeline}. 

$\mathcal{F}(\cdot)$ is normalized cross product operator on position difference:
\begin{equation}
\begin{aligned}
\mathcal{F}(\mathcal{P}_{fine}) 
    &= \frac{conv_h(\mathcal{P}_{fine})\times conv_v(\mathcal{P}_{fine})}{||conv_h(\mathcal{P}_{fine})||\cdot ||conv_v(\mathcal{P}_{fine}||}\\
\end{aligned}
\end{equation}
We compute position difference via nearby horizontal and vertical 3 pixels in texture space, giving rise to convolution kernels of $[-0.5, 0, 0.5]$ and $[-0.5, 0, 0.5]^\top$ for $conv_h$, $conv_v$ respectively. 

\begin{figure*}[t]
\begin{center}
   \includegraphics[width=1\linewidth]{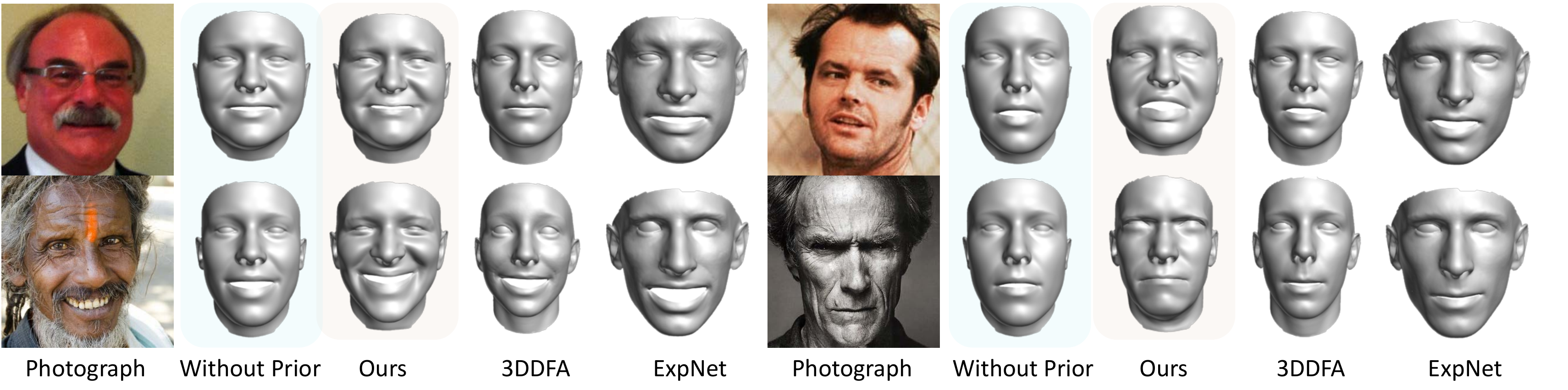}
\end{center}
   \caption{Comparisons of our emotion-driven proxy estimation vs. the state-of-the-art (3DDFA \cite{zhu2016face} and ExpNet \cite{chang2018expnet})}.
\label{fig:expression refine}
\end{figure*}

To reconstruct the appearance loss, we assume a Lambertian skin reflectance model and represent the global illumination using spherical harmonics (SH) to estimate environment lighting $S$ and albedo $\mathcal{I}_{albedo}$. Under this model, we can compute 
%the radiance $\mathcal{I}_{recon}$ and
the appearance loss $\mathcal{L}_{recon}$ as:
\begin{equation}
\begin{aligned}
\label{eq:rendering}
\mathcal{I}_{recon} &= \mathcal{I}_{albedo} \odot S(\mathcal{N}_{fine})\\
\mathcal{L}_{recon} &= ||\mathcal{I}_{input} - \mathcal{I}_{recon}||_1\\
\end{aligned}
\end{equation}

In order to back propagate $\mathcal{L}_{recon}$ to update displacement $G(u,v)$, we estimate albedo map $\mathcal{I}_{albedo}$ and environment lighting $S$ from sparse proxy vertices. Please refer to Section 1 of our supplementary material for detailed algorithm.

For training, we use high resolution facial images from the emotion dataset \textsl{AffectNet}~\cite{mollahosseini2017affectnet}, which contains more than 1M facial images collected from the Internet.

In our experiments, We also use HSV color space instead of RGB to accommodate environment lighting variations and employ a two-step training approach, i.e., only back propagate the loss of \textit{PDIM} for the first 10 epochs. We find the loss decreases much faster than starting with all losses. Moreover, we train 250 epochs for each facial area.
%based on our observation that the loss is smaller when there are more epochs, but the mesh is noisier in the experiment. 
To sum up, our expression estimation and detail synthesis networks borrow the idea of residual learning, breaking down the final target into a few small tasks, which facilitates training and improves performance in our tasks.

%\section{Data Preparation }
%In order to train our DFDG, we acquired a set of high-quality 3D facial scans based on structure light facial capture system \cite{cao2017sparse,stratou2011effect} and a large number of images for emotion net training and  non-supervised details synthesis. Our 3D scans consist of a total of 366 scans captured from 122 actors under structure lighting capture environment (both Polarized/non-Polarized), with each actor performing one natural expression and two casual expressions, as shown in figXXX. The global facial dataset obtains XXX high resolution website images of different ages, environment lighting, skin albedo et al. Our training data procession scheme includes following two parts: lighting and albedo estimation for global images, and detail extraction for facial scans.

\section{Experimental Results}
In order to verify the robustness of our algorithm, we have tested our emotion-driven proxy generation and facial detail synthesis approach on over 20,000 images (see supplementary material for many of these results).

~\\
\textbf{Expression Generation.} We downsample all images from $\textsl{AffectNet}$ dataset into $48\times48$ (the downsampling is only for proxy generation, not for detail synthesis) and use the Adam optimization framework with a momentum of 0.9. We train a total of 20 epochs and set learning rate to be $0.001$. Our trained \textsl{Emotion-Net} achieves a test accuracy of 52.2\%. Recall that facial emotion classification is a challenging task and even human annotators achieve only 60.7\% accuracy. Since our goal focuses on producing more realistic 3D facial models, we find this accuracy is sufficient for producing reasonable expression prior. 
% This large ambiguity inspires us to look up priors from the dictionary by using emotion feature similarity instead of resolving a mapping from image to expression parameters.

Fig. \ref{fig:expression refine}
shows some samples of our proxy generation results (without detail synthesis). Compared with the state-of-the-art solutions of 3D expression prediction \cite{zhu2016face,chang2018expnet}, we find that all methods are able to produce reasonable results in terms of eyes and mouth shape. However, the results from 3DDFA \cite{zhu2016face} and ExpNet \cite{chang2018expnet} exhibit less similarity with input images on regions such as cheeks, nasolabial folds and under eye bags while ours show significantly better similarity and depict person-specific characteristics. This is because such regions are not covered by facial landmarks. Using landmarks alone falls into the ambiguity mentioned in Section \ref{sec:emotion} and cannot faithfully reconstruct expressions on these regions. Our emotion-based expression predictor exploits global information from images and is able to more accurately capture expressions, especially for jowls and eye bags.

% fit the shape of mouth and eyes perfectly. However, as mentioned before, even similar facial landmark features may have vastly different expressions, as can be seen in Fig.\ref{fig:expression refine}. Our emotion based expression predictor shows better performance on emotions with higher accuracy and solves coherence between the shape of mouth/eyes and facial muscles, especially in the shape of the jowls and eye bags.

\begin{figure}[t]
\begin{center}
   \includegraphics[width=1\linewidth]{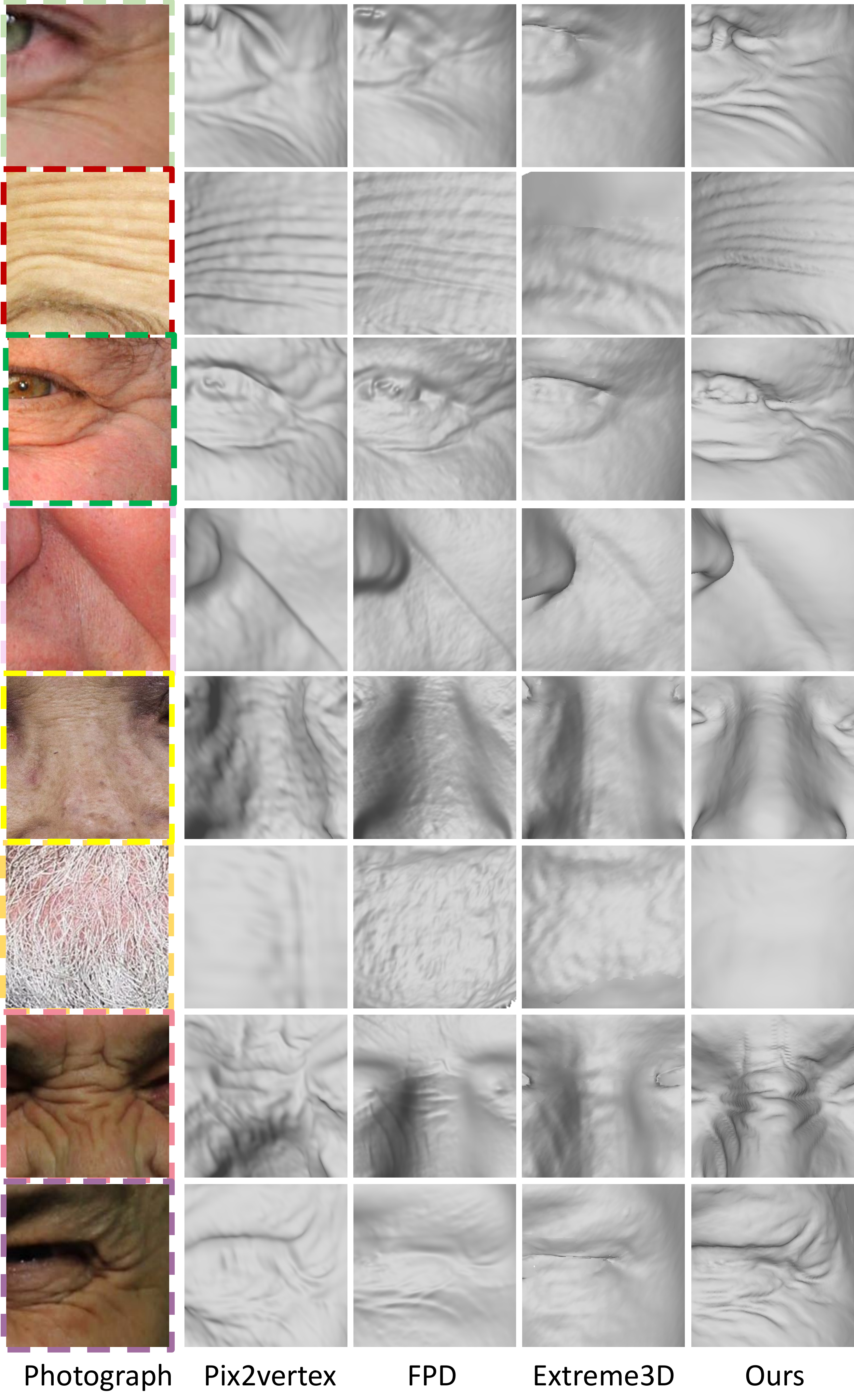}
\end{center}
\caption{Close-up views of the synthesized meshes using Pix2vertex~\cite{sela2017unrestricted}, FPD~\cite{li2018face}, Extreme3D~\cite{tran2018extreme} and ours.}
\label{fig:closeup} 
\end{figure}

~\\
\textbf{Facial Detail Synthesis.}
We sample a total of 10K patches for supervised training and 12K for unsupervised training. We train 250 epochs in total, and uniformly reduce learning rate from $0.0001$ to $0$ starting at 100th epoch. Note, we use supervised geometry loss for the first 15 epochs, and then alternate between supervised geometry loss and unsupervised appearance loss for the rest epochs.

Our facial detail synthesis aims to reproduce details from images as realistically as possible. Most existing detail synthesis approaches only rely on illumination and reflectance model~\cite{li2018face,tran2018extreme}. A major drawback of these methods lies in that their synthesized details resemble general object surface without considering skin's spatial correlation, as shown in close-up views in Fig.\ref{fig:closeup} (full mesh in supplementary material). Our wrinkles are more similar to real skin surface while the other three approaches are more like cutting with a knife on the surface. We attribute this improvement to combining illumination model with human face statistics from real facial dataset and wrinkle PCA templates.
%those in the lighting model and fail to consider skin's association, e.t. their bump looks like object surface instead of skin, as is shown in Fig. \ref{fig:expression refine} and close-up views in Fig.\ref{fig:closeup}(a full model in supplement). Our wrinkles are more similar to real skin surface while the other three approaches are more like cutting with a knife on the surface.

Our approach also has better performance on handling the surface noise from eyebrows and beards while preserving skin details (2 and 6th row of Fig.~\ref{fig:closeup}).
%To make a mesh comparison, we subdivide proxy mesh to about 46k vertices and per-vertices displace its position.
% Fig. \ref{fig:closeup} shows our approach has better permanence on handling the surface noise from eyebrows, braids, beard and our network can remove those obstruction and keep the skin details (2-4th of Fig.\ref{fig:closeup}).

\begin{figure}[t]
\begin{center}
  \includegraphics[width=1\linewidth]{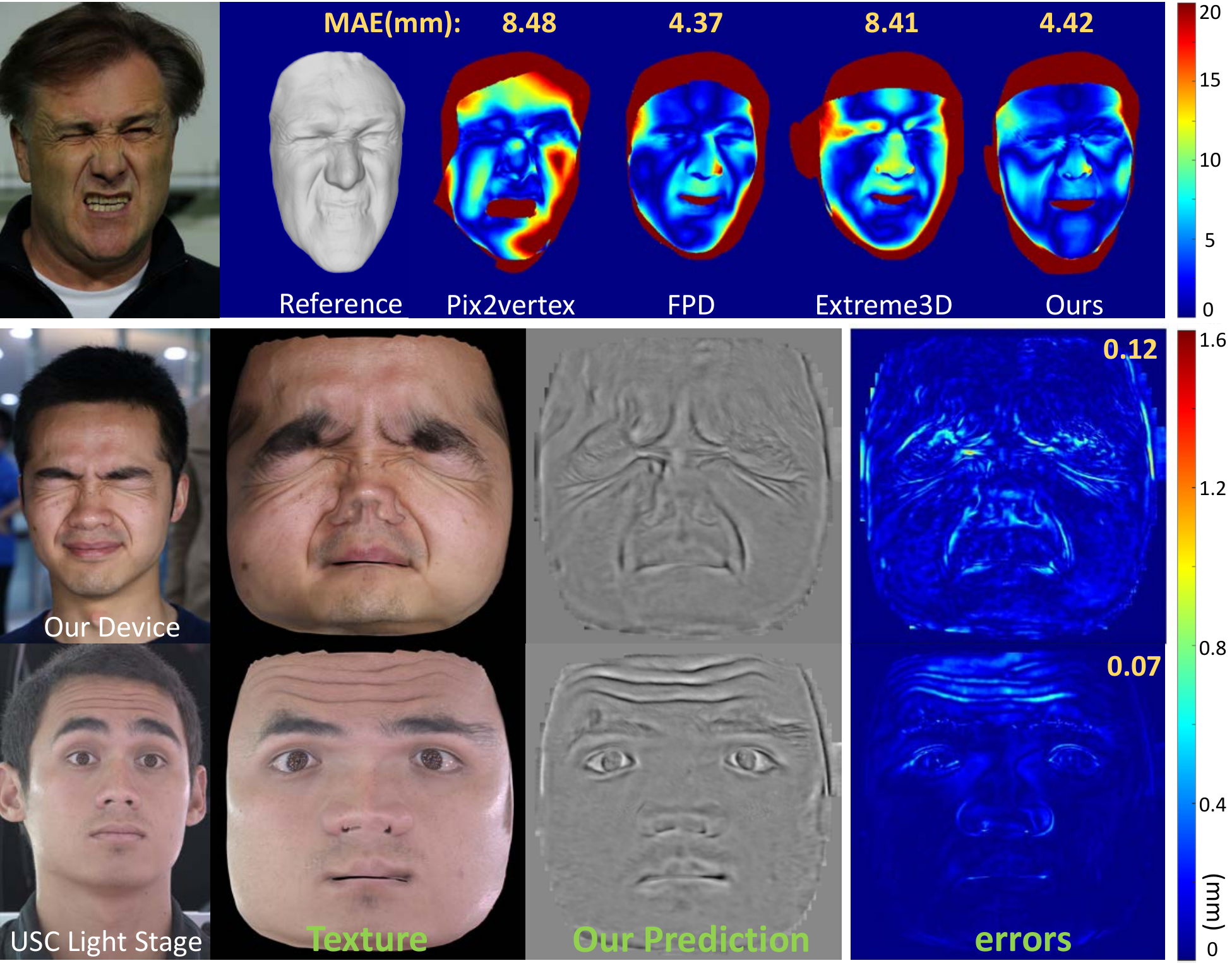}
\end{center}
\caption{Quantitative comparisons. Top row: the quantitative error maps (proxy + displacement) of Fig. \ref{fig:closeup} using different methods. Our approach achieves comparable performance on medium to large scale geometry (proxy) but produces much lower error on fine details such as the forehead and nasolabial folds. Two bottom rows: the error map of only the displacement on samples of our capture system and USC LightStage~\cite{ma2007rapid}.}
\label{fig:Quantitative}
\end{figure}

~\\
\textbf{Quantitative evaluation.}
We also carry out quantitative comparisons on both proxy mesh and displacement map. Overall, we observe our approach produces much lower errors on strong expressions, especially near the nose and eyebrows where shape deformations are strong. Fig.~\ref{fig:Quantitative} shows some sample results of proxy + displacement and displacement only errors.

Finally, our output displacement map is easy to integrate with existing rendering pipelines and can produce high-fidelity results, as shown in Fig.~\ref{fig:teaser}. 

\section{Conclusion and Future Work}
We have presented a single-image 3D face synthesis technique that can handle challenging facial expressions while preserving fine geometric structures. Our technique combines cues provided by emotion, expression, appearance, and lighting for producing high fidelity proxy geometry and fine geometric details. Specifically, we have conducted emotion prediction to obtain an expression-informed proxy and we have demonstrated that our approach can handle a wide range of expressions. For detail synthesis, our \textsl{Deep Facial Detail Net} (\textsl{DFDN}) employs both geometry and appearance loss functions and is trained on real data both captured by our system and from in-the-wild images. Comprehensive experiments have shown that our technique can produce, from a single image, ultra high quality 3D faces with fine geometric details under various expressions and lighting conditions. 

Although our solution is capable of handling a variety of lighting conditions, it has not yet considered the effects caused by occlusions (e.g., hair or glasses), hard shadows that may cause incorrect displacement estimations. For shadows, it may be possible to directly use the proxy to first obtain an ambient occlusion map and then correct the image. Shadow detection itself can be directly integrated into our learning-based framework with new sets of training data. Another limitation of our technique is that it cannot tackle low resolution images: our geometric detail prediction scheme relies heavily on reliable pixel appearance distribution. Two specific types of solutions we plan to investigate are to conduct (learning-based) facial image super-resolution that already accounts for lighting and geometric details as our input and to design a new type of proxy face model that includes deformable geometric details.

{\small
\bibliographystyle{ieee_fullname}
\bibliography{egbib}
}

\end{document}